\documentclass{article}
\usepackage{ijcai22}

\usepackage{times}
\usepackage{soul}
\usepackage{url}
\usepackage[hidelinks,colorlinks,citecolor=gray,linkcolor=gray]{hyperref}
\usepackage[utf8]{inputenc}
\usepackage[small]{caption}
\usepackage{graphicx}
\usepackage{amsmath}
\usepackage{amsthm}
\usepackage{booktabs}
\usepackage{algorithm}
\usepackage{algorithmic}

\usepackage{url}            
\usepackage{booktabs}       
\usepackage{amsfonts}       
\usepackage{nicefrac}       
\usepackage{microtype}      
\usepackage{xcolor}         

\usepackage{algorithm}
\usepackage{algorithmic}
\usepackage{graphicx}
\usepackage{textcomp}
\usepackage{xcolor}
\usepackage{subcaption}

\usepackage{color,xcolor}
\usepackage{epsfig}
\usepackage{graphicx}

\usepackage{adjustbox}
\usepackage{array}
\usepackage{booktabs}
\usepackage{multirow}

\usepackage{amsmath,amsfonts,amsthm,amssymb}
\usepackage{bm}
\usepackage{nicefrac}
\usepackage{microtype}

\usepackage{changepage}
\usepackage{extramarks}
\usepackage{fancyhdr}
\usepackage{lastpage}
\usepackage{setspace}
\usepackage{soul}
\usepackage{xspace}

\usepackage{comment}
\usepackage{enumitem}
\usepackage{chngcntr}

\title{Decentralized Unsupervised Learning of Visual Representations}

\author{
Yawen Wu$^1$ \and
Zhepeng Wang$^2$ \and
Dewen Zeng$^3$ \and
Meng Li$^4$ \and
Yiyu Shi$^3$ \And
Jingtong Hu$^1$\footnote{Corresponding author}
\affiliations
$^1$University of Pittsburgh\\
$^2$George Mason University\\
$^3$University of Notre Dame\\
$^4$Facebook
\emails
\{yawen.wu, jthu\}@pitt.edu,
zwang48@gmu.edu,
\{dzeng2, yshi4\}@pitt.edu,
meng.li@fb.com
}

\begin{document}



\maketitle

\begin{abstract}
Collaborative learning enables distributed clients to learn a shared model for prediction while keeping the training data local on each client.
However, existing collaborative learning methods require fully-labeled data for training, which is inconvenient or sometimes infeasible to obtain due to the high labeling cost and the requirement of expertise. 
The lack of labels makes collaborative learning impractical in many realistic settings.
Self-supervised learning can address this challenge by learning from unlabeled data.
Contrastive learning (CL), a self-supervised learning approach, can effectively learn visual representations from unlabeled image data. 
However, the distributed data collected on clients are usually not independent and identically distributed (non-IID) among clients, and each client may only have few classes of data, which degrades the performance of CL and learned representations.
To tackle this problem, we propose a collaborative contrastive learning framework consisting of two approaches: feature fusion and neighborhood matching, 
by which a unified feature space among clients is learned for better data representations.
Feature fusion provides remote features as accurate contrastive information to each client for better local learning. 
Neighborhood matching further aligns each client's local features to the remote features such that well-clustered features among clients can be learned.
Extensive experiments show the effectiveness of the proposed framework.
It outperforms other methods by 11\% on IID data and matches the performance of centralized learning.
\end{abstract}

\section{Introduction}
Collaborative learning is an effective approach for multiple distributed clients to collaboratively learn a shared model from decentralized data. 
In the learning process, each client updates the local model by using local data, and then a central server aggregates the local models to obtain a shared model.
In this way, collaborative learning enables learning from decentralized data \cite{mcmahan2017communication}
while keeping data local for privacy.
Collaborative learning can be applied to healthcare applications, where many personal devices such as mobile phones collaboratively learn to provide early warnings to cognitive diseases such as Parkinson’s and to assess mental health \cite{chen2020fedhealth}.
Collaborative learning can also be used for robotics, in which multiple robots learn a shared navigation scheme to adapt to new environments \cite{liu2019lifelong}. Compared with local learning, collaborative learning improves navigation accuracy by utilizing knowledge from other robots.

Existing collaborative learning approaches assume local data is fully labeled so that supervised learning can be used for the model update on each client.
However, labeling all the data is usually unrealistic due to high labor costs and the requirement of expert knowledge.
For example, in medical diagnosis, even if the patients are willing to spend time on labeling all the local data, the deficiency of expert knowledge of these patients will result in large label noise and thus inaccurate learned model.
The deficiency of labels makes supervised collaborative learning impractical.
Self-supervised learning can address this challenge by pre-training a neural network encoder with unlabeled data, followed by fine-tuning for a downstream task with limited labels.
Contrastive learning (CL), an effective self-supervised learning approach \cite{chen2020simple}, can learn data representations from unlabeled data to improve the model.
By integrating CL into collaborative learning, clients can collaboratively learn data representations by using a large amount of data without labeling.

In collaborative learning, data collected on clients are inherently far from IID \cite{hsu2020federated}, which results in two unique challenges when integrating collaborative learning with CL as collaborative contrastive learning (CCL) to learn high-quality representations.
The \emph{first challenge} is that 
each client only has a small amount of unlabeled data with limited diversity,
which prevents effective contrastive learning.
More specifically, 
compared with the global data (the concatenation of local data from all clients), each client only has a subset of the global data with a limited number of classes \cite{mcmahan2017communication,zhao2018federated,wu2021federated}.
For instance, in real-world datasets \cite{luo2019real}, each client only has one or two classes out of seven object classes.
Since conventional contrastive learning frameworks \cite{he2020momentum,chen2020simple} are designed for centralized learning on large-scale datasets with sufficient data diversity, directly applying them to local learning on each client will result in the low quality of learned representations.

The \emph{second challenge} is that each client focuses on learning its local data without considering the data on the other clients.
As a result, the features of data in the same class but from different clients may not be well-clustered even though they could have been clustered for improved representations.
Data are decentralized in collaborative learning and even if two clients have data of the same class, they are unaware of this fact and cannot leverage it to collaboratively learn to cluster these data.
Besides, even if one client has knowledge of other's data, since no labels are available, there is no easy way to identify the correct data clusters and perform clustering for better representations.

To address these challenges, we propose a collaborative contrastive learning (CCL) framework to learn visual representations from decentralized unlabeled data on distributed clients.
The framework employs contrastive learning \cite{he2020momentum} for local learning on each client and consists of two approaches to learn high-quality representations. 
The first approach is feature fusion and it provides remote features as accurate contrastive information to each client for better local learning.
To protect the privacy of remote features against malicious clients, we employ an encryption method \cite{huang2020instahide} to encrypt the images before generating their features. 
The second approach is neighborhood matching and it further aligns each client's local features to the fused features such that well-clustered features among clients are learned.

In summary, the main contributions of the paper include:

\begin{itemize}[itemsep=0pt,topsep=0pt,leftmargin=12pt]
	\item \textbf{Collaborative contrastive learning framework.} 
	We propose a framework with two approaches to
	learning visual representations from unlabeled data on distributed clients.
	The first approach improves the local representation learning on each client with limited data diversity, and the second approach further learns unified global representations among clients.
	\item \textbf{Feature fusion for better local representations.}
	We propose a feature fusion approach to leverage remote features 
	for better local learning while avoiding raw data sharing. 
	The remote features serve as negatives in the local contrastive loss to achieve a more accurate contrast with fewer false negatives and more diverse negatives. 
	\item \textbf{Neighborhood matching for improved global representations.}
	We propose a neighborhood matching approach to cluster decentralized data
	across clients.
	During local learning, each client identifies the remote features to cluster local data with and performs clustering.
	In this way, well-clustered features among clients can be learned.
\end{itemize}

\section{Background and Related Work}\label{sec:background_and_related}

\textbf{Contrastive Learning.} 
Contrastive learning is a self-supervised approach to learn an encoder (i.e. a convolutional neural network without the final classifier) for extracting visual representation vectors from the unlabeled input images by performing a proxy task of instance discrimination \cite{chen2020simple,he2020momentum,wu2018unsupervised,wu2021enabling}.
For an input image $x$, its representation vector $z$ is obtained by $z=f(x),\ z\in \mathbb{R}^d$, 
where $f(\cdot)$ is the encoder.
Let the representation vectors \emph{query} $q$ and \emph{key} $k^{+}$ form a positive pair, which are the representation vectors from two transformations (e.g. cropping and flipping) of the same input image.
Let $Q$ be the \emph{memory bank} with $K$ representation vectors stored, serving as negatives.
The positive pair \emph{query} $q$ and \emph{key} $k^{+}$ will be contrasted with each vector $n \in Q$ (i.e. negatives) by the loss function:
\begin{equation}\label{equ:mocoloss}
\ell_{q}=-\log \frac{\exp (q \cdot k^{+} / \tau)}{\exp (q \cdot k^{+} / \tau) + \sum_{n\in Q} \exp (q \cdot n / \tau)}
\end{equation}
Minimizing the loss will learn an encoder to generate visual representations. Then a classifier can be trained on top of the encoder by using limited labeled data.

However, existing contrastive learning approaches are designed for centralized learning \cite{chen2020simple,he2020momentum} and require sufficient data diversity for learning. When applied to each client in collaborative learning with limited data diversity, their performance will greatly degrade.
Therefore, an approach to increase the local data diversity on each client while protecting the shared information is needed.

\textbf{Collaborative Learning.}
The goal of collaborative learning is to learn a shared model by aggregating locally updated models from clients while keeping raw data on local clients \cite{mcmahan2017communication}. 
In collaborative learning, there are $C$ clients indexed by $c$. 
The training data $D$ is distributed among clients, and each client $c$ has a subset of the training data $D_c \subset D$.
There are recent works aiming to optimize the aggregation process \cite{reisizadeh2020fedpaq,nguyen2020fast}.
While our work can be combined with these works, for simplicity, we employ a typical collaborative learning algorithm \cite{mcmahan2017communication}. The learning is performed round-by-round.
In communication round $t$, the server randomly selects $\beta \cdot C$ clients $C^t$ and send them the global model with parameters $\theta^t$, where $\beta$ is the percentage of active clients per round.
Each client $c\in C^t$ updates the local parameters $\theta_c^t$ on local dataset $D_c$ for $E$ epochs to get $\theta_c^{t+1}$ by minimizing the loss $\ell_c(D_c,\theta^t)$. 
Then the local models are aggregated into the global model by averaging the weights 
$\theta^{t+1} \leftarrow \sum_{c\in C^t}\frac{|D_c|}{\sum_{i\in C^t} |D_i|}{\theta_c^{t+1}}$.
This learning process continues until the global model converges.

To improve the performance of collaborative learning on non-IID data, \cite{zhao2018federated,jeong2018communication} share local raw data (e.g.  images) among clients.
However, sharing raw data among clients will cause privacy concerns \cite{li2018federated}.
Besides, they need fully labeled data to perform collaborative learning,
which requires expert knowledge and potentially high labeling costs.
Therefore, an approach to performing collaborative learning with limited labels and avoiding sharing raw data is needed.

\begin{figure*}[!htb]
	\centering
	\includegraphics[width=\linewidth]{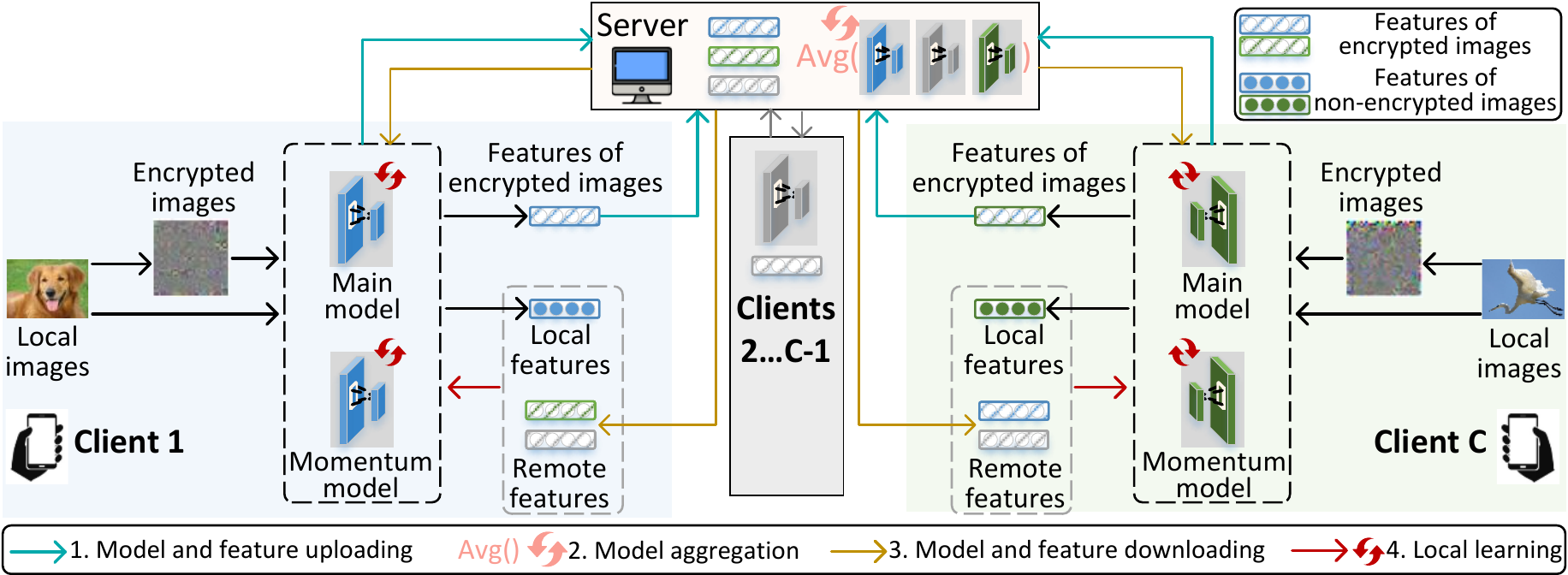}
	\caption{Overview of the proposed collaborative contrastive learning (CCL) framework. Four steps are performed in each learning round, including model and feature uploading from clients to the server, model aggregation on the server, model and feature downloading to clients, and local learning.
	Local learning is performed by the proposed feature fusion in Sect. \ref{sec:feature_exchange} and neighborhood matching in Sect. \ref{sec:neigh_match}.
	}
	\label{fig:overview}
\end{figure*}

\section{Framework Overview}\label{sec:}

We propose a collaborative contrastive learning (CCL) framework to learn representations from unlabeled data on distributed clients.
These distributed data cannot be combined in a single location to construct a centralized dataset due to privacy and legal constraints \cite{kairouz2019advances}.
The overview of the proposed framework is shown in Figure \ref{fig:overview}. 
There is a central server that coordinates multiple clients to learn representations.
The local model on each client is based on MoCo \cite{he2020momentum}.
CCL follows the proposed feature fusion technique to reduce the false negative ratio on each client for better local learning (Sect. \ref{sec:feature_exchange}). 
Besides, based on fused features, CCL further uses the proposed neighborhood matching to cluster representations of data from different clients to learn unified representations among clients (Sect. \ref{sec:neigh_match}).

Before introducing the details of feature fusion and neighborhood matching, we present the proposed CCL process. 
CCL is performed round-by-round and there are four steps in each round as shown in Figure \ref{fig:overview}. 
First, each client $c$ uploads its latest model (consisting of the main model $f_q^c$ and momentum model $f_k^c$) and latest features $\overline{Q}_{l,c}$ of encrypted images to the server.
Second, the server aggregates main models from clients by $\theta_q \leftarrow \sum_{c\in C}\frac{|D_c|}{|D|}{\theta_q^{c}}$ and momentum models by $\theta_k \leftarrow \sum_{c\in C}\frac{|D_c|}{|D|}{\theta_k^{c}}$ to get updated $f_q$ and $f_k$, where $|D_c|$ is the data size of client $c$ and $|D|$ is the total data size of $|C|$ clients. The server also combines features as $\overline{Q} = \{\overline{Q}_{l,c}\}_{c \in C}$.
Third, the server downloads the aggregated models $f_q$ and $f_k$ and combined features $\overline{Q}$ excluding $\overline{Q}_{l,c}$ as $Q_{s,c} = \{ \overline{Q} \  \char`\\ \ \overline{Q}_c \}$ to each client $c$.
Fourth, each client updates its local models with $f_q$ and $f_k$, and then performs local contrastive learning for multiple epochs with local features $Q_{l,c}$ and remote features $Q_{s,c}$ by using loss Eq.(\ref{equ:final_loss}) including contrastive loss with fused features Eq.(\ref{equ:contrast_batch}) and neighborhood matching Eq.(\ref{equ:neighbor_matching}). 
During local contrastive learning, to generate features $\overline{Q}_{l,c}$ for uploading in the next round, images $x$ are encrypted by \emph{InstaHide} \cite{huang2020instahide} as $\tilde{x}$ and fed into momentum model $f_k^c$. In this way, even if a malicious client can ideally recover $\tilde{x}$ from the features $\overline{Q}_{l,c}$, which is already very unlikely in practice, it still cannot reconstruct $x$ from $\tilde{x}$ since \emph{Instahide} effectively hides information contained in $x$.
Next, we present the details of local contrastive learning, including feature fusion to reduce false negative ratio in Sect. \ref{sec:feature_exchange} and neighborhood matching for unified representations in Sect. \ref{sec:neigh_match}.

\section{Local Learning with Feature Fusion}\label{sec:feature_exchange}

Next, we focus on how to perform local CL in each round of CCL. 
We first present the key challenge of CL on each client, which does not exist in conventional centralized CL.
Then we propose feature fusion to tackle this challenge and introduce how to perform local CL with fused features.

\textbf{Key challenge:} \emph{Limited data diversity} causes a high false negative (FN) ratio on each client.
A low FN ratio is crucial to achieving accurate CL \cite{kalantidis2020hard}.
For one image sample $q$, FNs are features that we use as negative features but actually correspond to images of the same class as $q$.
In centralized CL, the percentage of FNs is inherently low since diverse data are available.
The model has access to the whole dataset $D$ with data from all the classes instead of a subset $D_c$ as in collaborative learning.
Thus, when we randomly sample negatives from $D$, the FN ratio is low.  
For instance, when dataset $D$ has 1000 balanced classes and the negatives $n$ are randomly sampled, for any image $q$ to be learned, only $\frac{1}{1000}$ of $n$ are from the same class as $q$ and are FNs.

However, in collaborative learning, the FN ratio is inherently high on each client due to the limited data diversity, which significantly degrades the performance of contrastive learning.
For instance, in real-world datasets \cite{luo2019real}, one client can have only one or two classes out of seven classes.
With limited data diversity on each client, when learning image sample $q$, many negatives $n$ to contrast with will be from the same class as $q$ and are FNs.
To perform contrastive learning by minimizing the contrastive loss in Eq.(\ref{equ:mocoloss}), the model scatters the FNs away from $q$, which should have been clustered since they are from the same class.
As a result, the representations of samples from the same class will be scattered instead of clustered and degrade the learned representations.

\subsection{Feature Fusion}\label{sec:feature_fusion}

To address this challenge, we propose feature fusion to share negatives in the feature space (i.e. the output vector of the encoder), which reduces FN and improves the diversity of negatives while avoiding raw data sharing.
Let $Q_{l,c}$ be the memory bank of size $K$ for local features of non-encrypted images on client $c$, and let $\overline{Q}_{l,c}$ be features of encrypted images. 
In one round $t$ of CCL, features $\overline{Q}_{l,c}$ of encrypted images on each client $c$ will be uploaded to the server (i.e. step 1 in Figure \ref{fig:overview}).
The server also downloads combined features $\overline{Q}$ excluding $\overline{Q}_{l,c}$ to each client $c$ (i.e. step 3 in Figure \ref{fig:overview})
to form its memory bank of remote negatives $Q_{s,c}$ as follows. 
\begin{equation}\label{equ:shared_negatives}
Q_{s,c} = \{\overline{Q}_{l,i}\ |\ 1 \le i \le |C|, i\ne c \}
\end{equation}
where $C$ is the set of all clients.

On client $c$, with local negatives $Q_{l,c}$ and remote negatives $Q_{s,c}$, the loss for sample $q$ is defined as:
\begin{equation}\label{equ:moco_loss_shared_negative}
\ell_{q}= -\log \left[ \frac{\exp (q \cdot k^{+} / \tau)}{\exp (q \cdot k^{+} / \tau) + \sum_{n\in \{Q_{l} \cup Q_{s}\}} \exp (q \cdot n / \tau) } \right]
\end{equation}
where we leave out the client index $c$ in $Q_{l,c}$ and $Q_{s,c}$ for conciseness.
$\ell_{q}$ is the negative log-likelihood over the probability distribution generated by applying a softmax function to a pair of input $q$ and its positive $k^{+}$, negatives $n$ from both local negatives $Q_{l}$ and remote negatives $Q_{s}$.

\textbf{Effectiveness of feature fusion.}
The remote negatives $Q_{s}$ reduce the FN ratio in local contrastive learning and improve the quality of learned representations on each client.
More specifically, in collaborative learning with non-IID data, 
we assume the global dataset $D$ has $M$ classes of data, each class with the same number of data. Each client $c \in C$ has a subset $D_c \subset D$ of the same length in $m$ classes ($m \le M$) \cite{zhao2018federated,mcmahan2017communication}. 
For a sample $q$ on client $c$, when only local negatives $Q_{l,c}$ are used, $\frac{1}{m}|Q_{l,c}|$ negatives will be in the same class as $q$, which results in an FN ratio $R_{FN}=\frac{1}{m}$. 
Since $m$ is usually small (e.g. 2) due to limited data diversity, the FN ratio $R_{FN}$ will be large (e.g. 50\%) and degrade the quality of learned representations.
Different from this, when remote negatives are used, 
the FN ratio is:
\begin{equation}\label{equ:fp_with_share}
R_{FN}(q)= \frac{\frac{1}{m}|Q_{l,c}| + \sum_{i\in C, i\ne c}\mathbb{I}(i,q) \frac{1}{m}|Q_{l,i}|}{|Q_{l,c}|+\sum_{i\in C, i\ne c}|Q_{l,i}|} \le \frac{1}{m}
\end{equation}
where $\mathbb{I}(i,q)$ is an indicator function that equals 1 when client $i$ has data of the same class as $q$, and 0 otherwise.

In most cases, $R_{FN}(q)$ is effectively reduced by the remote negatives.
\emph{First}, in the extreme case of non-IID data distribution, where the classes on each client are mutually exclusive \cite{zhao2018federated}, all $\mathbb{I}(i,q)$ equal 0 and $R_{FN}(q)=\frac{1}{m}\frac{|Q_{l,c}|}{|Q_{l,c}|+\sum_{i\in C, i\ne c}|Q_{l,i}|}=\frac{1}{m|C|}\ll\frac{1}{m}$.
With the remote negatives, the FN ratio is effectively reduced by a factor $|C|$.
\emph{Second}, as long as not all clients have data of the same class as $q$, some elements in $\{\mathbb{I}(i,q)\}_{i=1,\ i\ne c}^{|C|}$ will be 0, and $R_{FN}(q)$ in Eq.(\ref{equ:fp_with_share}) will be smaller than $\frac{1}{m}$. In this case, the FP ratio is also reduced.
\emph{Third}, even if the data on each client is IID and all $\mathbb{I}(i,q)$ equal 1, which is unlikely in realistic collaborative learning \cite{hsu2020federated}, the FN ratio $R_{FN}(q)$ will be $\frac{1}{m}$. In this case, while $R_{FN}(q)$ is the same as that without remote negatives, the increased diversity of negatives from other clients can still benefit the local contrastive learning.

\textbf{Further reducing the false negative ratio.}
To further reduce the FN ratio,
we propose to exclude the local negatives by removing $Q_l$ in the denominator of Eq.(\ref{equ:moco_loss_shared_negative}) and only keeping remote negatives $Q_s$.
The corresponding FN ratio becomes:
\begin{equation}\label{equ:fp_no_local}
R_{FN}^{\prime}(q)= \frac{\sum_{i\in C, i\ne c}\mathbb{I}(i,q) \frac{1}{m}|Q_{l,i}|}{\sum_{i\in C, i\ne c}|Q_{l,i}|}
\le R_{FN}(q)
\end{equation}
As long as not all other clients have data in the same class as $q$, some $\mathbb{I}(i,q)$ will be 0.
In this way, $R_{FN}^{\prime}(q) < R_{FN}(q)$ and the FN ratio is further reduced.
Based on the loss $\ell_{q}$ for one sample $q$ in 
Eq.(\ref{equ:moco_loss_shared_negative}),
the contrastive loss for one mini-batch $B$ is:
\begin{equation}\label{equ:contrast_batch}
\mathcal{L}_{contrast}=\frac{1}{|B|} \sum_{q \in B} {\ell_{q}}
\end{equation}

\section{Local Learning with Neighborhood Matching}\label{sec:neigh_match}

\begin{figure}[!htb]
	\centering
	\includegraphics[width=\columnwidth]{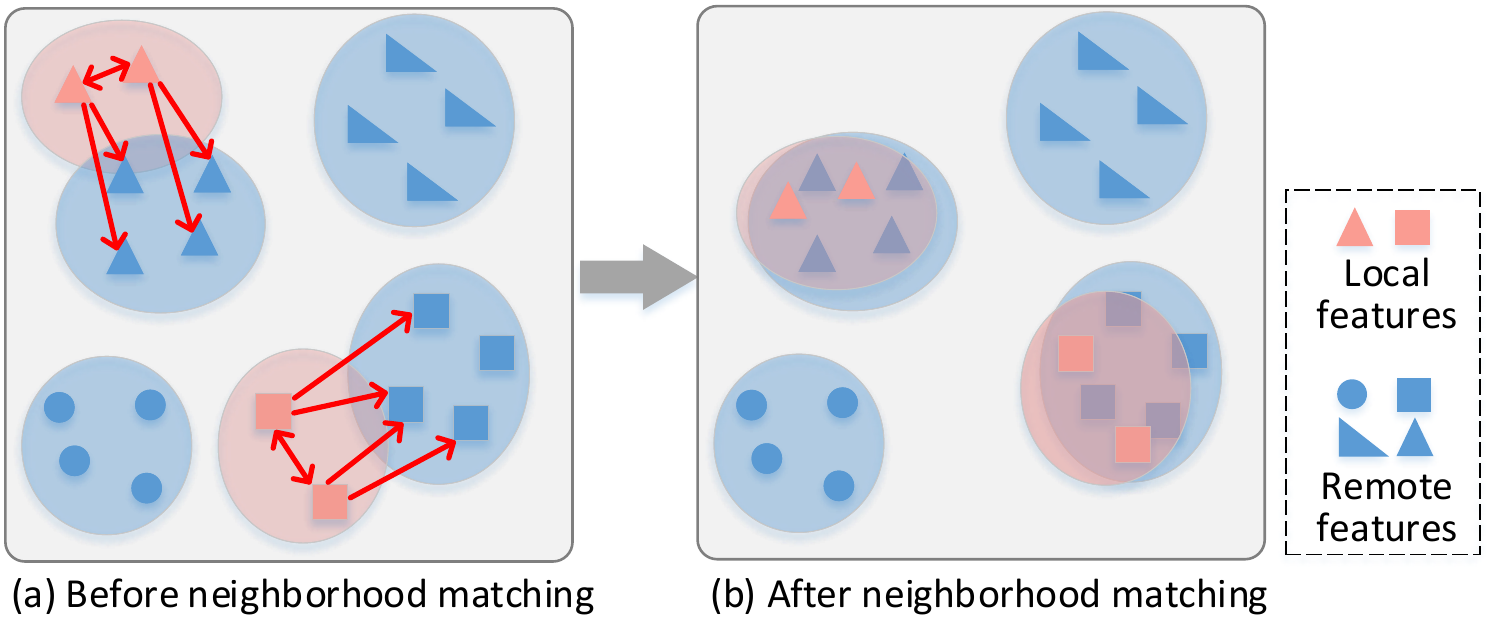}
	\caption{Neighborhood matching aligns each client's local features to the remote features such that well-clustered features among clients are learned.}
	\label{fig:neighborhood_matching}
\end{figure}

In local contrastive learning, each client focuses on learning its local data without considering data on the other clients.
As a result, the features of data in the same class but from different clients may not be well-clustered even though they could be clustered for improved representations.

\textbf{Challenge:} 
To cluster local features to the correct remote features, one has to
identify local data and remote features that are in the same class.
However, since no labels are available for local data and remote features, there is no easy way to identify the correct clusters to push local features to.

To address this challenge,
we propose a neighborhood matching approach to identify the remote features to cluster local data with and define an objective function to perform the clustering.
First, 
during local learning on one client, as shown in Figure \ref{fig:neighborhood_matching}, for each local sample we find $N$ nearest features from both the \textit{remote} and \textit{local} features as neighbors. Then the features of the local sample will be pushed to these neighbors by the proposed entropy-based loss. 
Since the model is synchronized from the server to clients in each communication round, the remote and local features are encoded by similar models on different clients.
Therefore, the neighbors are likely to be in the same class as the local sample being learned, and clustering them will improve the learned representations of global data. In this way, the global model is also improved when aggregating local models.

\textbf{Identifying neighbors.}
To push each local sample close to its neighbors, we minimize the entropy of one sample's matching probability distribution to either a remote feature or a local feature. To improve the robustness, we match one sample to $N$ nearest features at the same time, instead of only one nearest feature. 
By minimizing the entropy for $N$ nearest neighbors, the sample's matching probability to each of the nearest neighbors will be individually certain.

For each local sample $q_i$, we regard top-\textit{N} closest features, either from local or remote features, as neighbors. To find the neighbors, we first define the neighbor candidates as:
\begin{equation}\label{equ:neighbor_candidates}
Q^\prime=\left \{ Q_{s+l,i} |\ i \sim \mathcal{U}(|Q_s|+K,K) \right \} 
\end{equation}
where $i \sim \mathcal{U}(|Q_s|+K,K)$ samples $K$ integer indices from $[|Q_s|+K]$ randomly at uniform. 
$Q_{s+l,i}$ is the element with index \textit{i} in the union of remote and local features $Q_{s}\cup Q_{l}$. 
For one local sample $q_i$, the neighbors $P(q_i)$, which are the top-\textit{N} nearest neighbor candidates $Q^\prime$, is given by:
\begin{equation}\label{equ:topN_neighbors}
P(q_i) = \{ Q^{\prime}_j \ |\ j \in topN(S_{i,m}), 1 \le m \le K \}
\end{equation}
where $S_{i,m}=sim(q_i,n_m)={q_i^T \cdot n_m}/{\|q_i\|\|n_m\|}$ is the cosine similarity between $q_i$ and one neighbor candidate $n_m \in Q^\prime $.

\textbf{Neighborhood matching loss.}
To make the probability of $q_i$ matching to each $n_j \in P(q_i)$ individually certain, we consider the set: 
\begin{equation}
L_j = \{ n_j \} \cup \{ Q^\prime \  \char`\\ \ P(q_i) \} \in \mathbb{R}^{(K-N+1)\times d}
\end{equation}
where $d$ is the dimension of one feature vector. $L_j$ contains one of the top-$N$ nearest neighbors $n_j$ and neighbor candidates excluding all other top-$N$ nearest neighbors.

Given $L_j$,
the probability that sample $q_i$ is matched to one of the neighbors $n_a \in L_j$ is:
\begin{equation}
p_{i,j,a} = \frac{ \exp(q_i^T \cdot n_a / \tau_{nm} ) } { \sum_{n\in L_j} \exp(q_i^T \cdot n / \tau_{nm} ) },\ n_a \in L_j
\end{equation}
The temperature $\tau_{nm}$ controls the softness of the probability distribution \cite{hinton2015distilling}.
Since $n_j$ has the largest cosine similarity with $q_i$ for  $ n \in L_j$, $p_{i,j,a}$ will have the largest value when $n_a=n_j$ for $n_a \in L_j$.
In this way, when minimizing the entropy of the probability distribution $\{p_{i,j,a}\}_{n_a \in L_j}$, the matching probability of $q_i$ and $n_j$ will be maximized.

For one mini-batch $B$, to match each sample to its $N$ nearest neighbors, the entropy for all samples in this mini-batch is calculated as:
\begin{equation}\label{equ:neighbor_matching}
\mathcal{L}_{neigh} = -\frac{1}{|B|} \sum_{i\in B} \frac{1}{N} \sum_{j=1}^{N} \sum_{a=1}^{K-N+1} p_{i,j,a} \log (p_{i,j,a})
\end{equation}
where $K-N+1$ is the number of features in $L_j$, and $N$ is the number of nearest neighbors to match.
By minimizing $\mathcal{L}_{neigh}$, each $i\in B$ will be aligned to its top-$N$ nearest neighbors.

\textbf{Final loss.} Based on the contrastive loss with fused features in Eq.(\ref{equ:contrast_batch}) and neighborhood matching loss in Eq.(\ref{equ:neighbor_matching}), the overall objective is formulated as:
\begin{equation}\label{equ:final_loss}
\mathcal{L} = \mathcal{L}_{contrast} + \lambda \mathcal{L}_{neigh}
\end{equation}
where $\lambda$ is a weight parameter.

\section{Experimental Results}\label{sec:experiment}

\textbf{Datasets, model architecture, and distributed settings.}
We evaluate the proposed approaches on three datasets, including CIFAR-10 \cite{krizhevsky2009learning}, CIFAR-100 \cite{krizhevsky2009learning}, and Fashion-MNIST \cite{xiao2017fashion}. 
We use ResNet-18 as the base encoder and use a 2-layer MLP to project the representations to 128-dimensional feature space \cite{chen2020simple,he2020momentum}. 
For each of the three datasets, we consider one IID setting and two non-IID settings.
The detailed collaborative learning settings and training details can be found in the Appendix.

\textbf{Metrics.}
To evaluate the quality of learned representations, we use standard metrics for centralized self-supervised learning, including \textit{linear evaluation} and \textit{semi-supervised learning}  \cite{chen2020simple}. 
Besides, we evaluate by \textit{collaborative finetuning} for realistic collaborative learning.
In linear evaluation, a linear classifier is trained on top of the frozen base encoder, and the test accuracy represents the quality of learned representations. 
We first perform collaborative learning by the proposed approaches without labels to learn representation. Then we \textit{fix} the encoder and train a linear classifier on the 100\% labeled dataset on top of the encoder.
The classifier is trained for 100 epochs by the SGD optimizer following the hyper-parameters from \cite{he2020momentum}.
In semi-supervised learning, we first train the base encoder without labels in collaborative learning. Then we append a linear classifier to the encoder and \textit{finetune} the whole model on 10\% or 1\% labeled data for 20 epochs with SGD optimizer following the hyper-parameters from \cite{caron2020unsupervised}.
In collaborative finetuning, 
the learned encoder by the proposed approaches is used as the initialization for finetuning the whole model by supervised collaborative learning \cite{mcmahan2017communication} with few locally labeled data on clients.
Detailed collaborative finetuning settings can be found in the Appendix.

\textbf{Baselines.}
We compare the proposed methods with multiple approaches.
\emph{Predicting Rotation} is a self-supervised learning approach by predicting the rotation of images \cite{gidaris2018unsupervised}.
\emph{DeepCluster-v2} is the improved version of DeepCluster \cite{caron2020unsupervised,caron2018deep} and achieves SOTA performance.
\emph{SwAV} and \emph{SimCLR} are SOTA approaches for self-supervised learning \cite{caron2020unsupervised,chen2020simple}.
We combine these approaches with FedAvg as \emph{FedRot}, \emph{FedDC}, \emph{FedSwAV}, and \emph{FedSimCLR}.
\emph{FedCA} is the SOTA collaborative unsupervised learning approach with a shared dictionary and online knowledge distillation \cite{zhang2020federated}.
Besides, we compare with two methods as upper bounds.
\emph{MoCo} \cite{he2020momentum} is a centralized contrastive learning method assuming all data are combined in a single location. We compare with MoCo since the local model in the proposed methods is based on it.
\emph{FedAvg} \cite{mcmahan2017communication} is a fully supervised collaborative learning method.

\subsection{Linear Evaluation}

\begin{figure}[!htb]
	\centering
		\includegraphics[width=\linewidth]{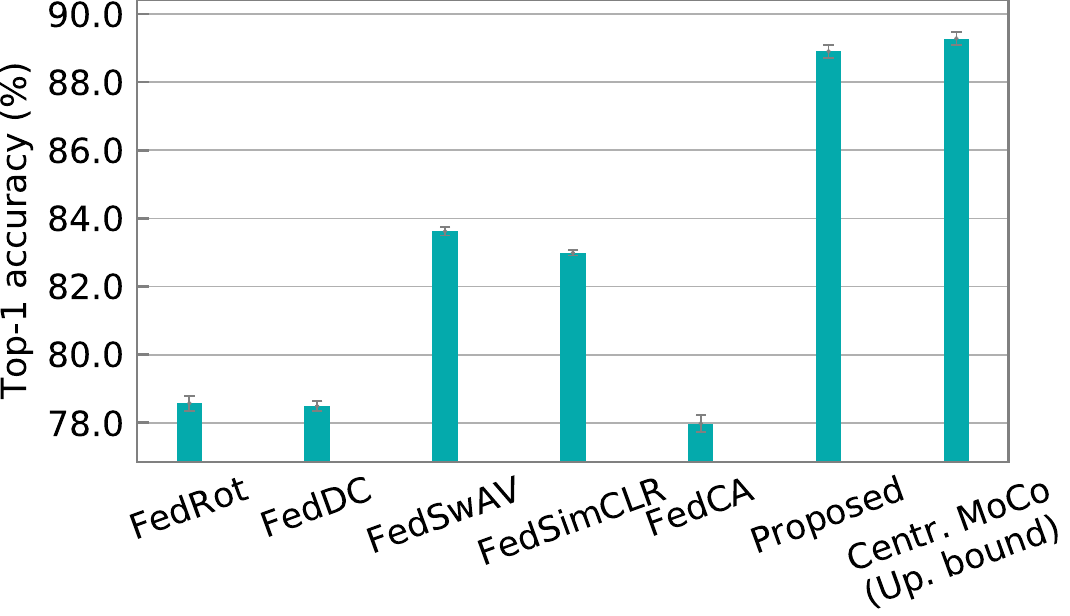}
    \caption{\textbf{Linear evaluation accuracy on CIFAR-10
    in IID setting.} The classifier is trained by 100\% labels on a \textit{fixed} encoder learned by different approaches. Error bar denotes standard derivation over three independent runs. 
    Centralized MoCo is the upper bound method.
	}
	\label{fig:exp_linaer_cifar10_fl1}
\end{figure}

\textbf{Linear evaluation on CIFAR-10.}
We evaluate the proposed approaches by linear evaluation with 100\% data labeled for training the classifier on top of the \textit{fixed} encoder learned with unlabeled data by different approaches. 
The proposed approaches significantly outperform other methods and even match the performance of the centralized upper bound method.
The results on CIFAR-10 
in the IID setting are shown in Figure \ref{fig:exp_linaer_cifar10_fl1}.
On CIFAR-10, the proposed approaches achieve 88.90\% top-1 accuracy, only 0.38\% below the upper bound method centralized MoCo.
The proposed approaches also outperform the SOTA method FedCA by +10.92\% top-1 accuracy and the best-performing baseline FedSwAV by +5.28\%.

\begin{table}[!htb]
	\centering
	\caption{\textbf{Linear evaluation on CIFAR-10, CIFAR-100, and FMNIST under the IID and two non-IID settings.} 100\% labeled data are used for learning the classifier on the \textit{fixed} encoder and top-1 accuracy is reported. 
	Centralized MoCo and Supervised FedAvg are the upper bound methods.
	}
	\label{tab:exp_linaer_all_settings_cifar10_cifar100_fmnist}
	\setlength\tabcolsep{2.0pt}
	\renewcommand{\arraystretch}{1.0}
	\resizebox{1.0\columnwidth}{!}{
		\begin{tabular}{lccccccccc}
			\toprule
			& \multicolumn{3}{c}{CIFAR-10}                     & \multicolumn{3}{c}{CIFAR-100}                    & \multicolumn{3}{c}{FMNIST}                       \\ \cmidrule(lr){2-4}\cmidrule(lr){5-7}\cmidrule(lr){8-10}
			Method     & IID            & Non-1      & Non-2      & IID            & Non-1      & Non-2      & IID            & Non-1      & Non-2      \\ \midrule
			FedRot     & 78.57          & \underline{75.88}          & 70.98          & 45.80          & 44.57          & 43.15          & 83.74          & 82.65          & 82.90          \\
			FedDC      & 78.49          & 69.97          & 69.34          & 49.27          & 49.06          & 47.21          & 88.41          & 85.92          & 88.35          \\
			FedSwAV    & \underline{83.62}          & 75.07          & 75.36          & \underline{55.51}          & \underline{51.45}          & \underline{53.77}          & \underline{89.63}          & \underline{87.11}          & \underline{89.75}          \\
			FedSimCLR  & 82.99          & 71.23          & 73.30          & 48.83          & 45.67          & 48.46          & 88.45          & 84.41          & 86.23          \\
			FedCA      & 77.98          & 75.57          & \underline{75.50}          & 48.93          & 47.70          & 48.22          & 86.98          & 86.22          & 86.46          \\
            Proposed   & \textbf{88.90} & \textbf{79.07} & \textbf{78.31} & \textbf{61.91} & \textbf{57.54} & \textbf{58.63} & \textbf{91.26} & \textbf{88.13} & \textbf{90.08} \\ \midrule
            \multicolumn{10}{l}{\textit{Upper bounds}} \\
            MoCo (Centralized) & 89.28 & --- & --- & 63.72 & --- & --- & 91.97 & --- & ---  \\
            FedAvg (Supervised) & 92.88 & 60.60 & 59.03 & 73.08 & 67.59 & 66.90 & 94.12 & 77.08 & 69.92  \\
			\bottomrule
	\end{tabular}}
\end{table}

\textbf{Linear evaluation on various datasets and distributed settings.}
We evaluate the proposed approaches on different datasets and collaborative learning settings. 
The results under the IID setting, non-IID settings 1 and 2 on CIFAR-100 and FMNIST are shown in Table \ref{tab:exp_linaer_all_settings_cifar10_cifar100_fmnist}.
The linear classifier is trained on top of the \textit{fixed} encoder learned with unlabeled data by different approaches.
Under all the three collaborative learning settings and on both datasets, the proposed approaches significantly outperform the baselines.

For example, on CIFAR-100 the proposed approaches outperform the best-performing baseline by 6.40\%, 6.09\%, and 4.86\% under three collaborative learning settings, respectively.
Besides, compared with the two upper bound methods, the proposed methods match the performance of the upper bound centralized MoCo under IID setting, and outperforms supervised FedAvg on CIFAR-10 under non-IID settings.

\subsection{Semi-Supervised Learning}

\begin{table}[h]
	\centering
	\caption{
		\textbf{Semi-supervised learning under IID setting (top) and non-IID setting 1 (bottom).} We \textit{finetune} the encoder and classifier with different ratios of labeled data and report the top-1 accuracy.
	}
	\label{tab:exp_semi_all_fl1_fl2}
	\begin{minipage}{\linewidth}
		\setlength\tabcolsep{3.0pt}
		\renewcommand{\arraystretch}{1.0}
		\resizebox{1.0\columnwidth}{!}{
			\begin{tabular}{lccccccccc}
				\toprule
				& \multicolumn{2}{c}{CIFAR-10}    & \multicolumn{2}{c}{CIFAR-100}   & \multicolumn{2}{c}{FMNIST}      \\ \cmidrule(lr){2-3}\cmidrule(lr){4-5}\cmidrule(lr){6-7}
				Labeled ratio & 10\%           & 1\%            & 10\%           & 1\%            & 10\%           & 1\%            \\ \midrule
				FedRot        & 85.38          & 71.62          & 43.78          & 19.84          & 91.23          & 47.94          \\
				FedDC         & 78.88          & 44.18          & 40.69          & 11.93          & 88.97          & 30.25          \\
				FedSwAV       & 84.51          & 48.96          & \underline{50.23}          & 13.82          & 90.48          & 62.08          \\
				FedSimCLR     & \underline{86.05}          & \underline{75.36}          & 49.54          & \underline{27.45}          & 91.28          & \underline{84.46}          \\
				FedCA          & 84.15          & 41.25          & 48.57          & 8.13           & \underline{91.67}          & 36.93          \\
				Proposed      & \textbf{89.27} & \textbf{84.79} & \textbf{58.49} & \textbf{40.71} & \textbf{92.18} & \textbf{85.63} \\ \midrule
				MoCo (Centralized) & 88.44 & 81.75 & 57.76 & 37.79 & 92.46 & 86.78 \\
				\bottomrule
		\end{tabular}}
	\end{minipage}
	\begin{minipage}{\linewidth}
		\setlength\tabcolsep{3.0pt}
		\renewcommand{\arraystretch}{1.0}
		\resizebox{1.0\columnwidth}{!}{
			\begin{tabular}{lccccccccc}
				\toprule
				& \multicolumn{2}{c}{CIFAR-10}    & \multicolumn{2}{c}{CIFAR-100}   & \multicolumn{2}{c}{FMNIST}      \\ \cmidrule(lr){2-3}\cmidrule(lr){4-5}\cmidrule(lr){6-7}
				Labeled ratio & 10\%           & 1\%            & 10\%           & 1\%            & 10\%           & 1\%            \\ \midrule
				FedRot        & 77.82 & 58.48          & 43.50          & 18.80          & \underline{90.80}          & 60.72          \\
				FedDC         & 71.25          & 31.85          & 40.85          & 11.42          & 86.80          & 37.91          \\
				FedSwAV       & 78.25          & 39.87          & 46.58          & 14.11          & 88.77          & 35.85          \\
				FedSimCLR     & 78.49          & 58.13          & 46.89          & \underline{23.86}          & 90.41          & \underline{80.72}          \\
				FedCA          & \underline{79.75}          & \underline{58.76}          & \underline{48.10}          & 8.07           & 89.60          & 38.98          \\
				Proposed      & \textbf{84.01}          & \textbf{67.87} & \textbf{54.85} & \textbf{31.29} & \textbf{91.29} & \textbf{82.13} \\ \midrule
				MoCo (Centralized) & 88.44 & 81.75 & 57.76 & 37.79 & 92.46 & 86.78 \\
				\bottomrule
		\end{tabular}}
	\end{minipage}
\end{table}

We further evaluate the proposed approaches by \textit{semi-supervised learning}, where both the encoder and classifier are \textit{finetuned} with 10\% or 1\% labeled data after learning the encoder on unlabeled data by different approaches.
We evaluate the approaches under the IID collaborative learning setting and two non-IID collaborative learning settings.
Table \ref{tab:exp_semi_all_fl1_fl2}
shows the comparison of our results against the baselines under the IID collaborative learning setting (top) and non-IID setting 1 (bottom).
Our approach significantly outperforms the self-supervised baselines with 10\% and 1\% labels.
Notably, the proposed methods even outperform the upper bound method centralized MoCo on CIFAR-10 and CIFAR-100 datasets under the IID setting.
Results under non-IID collaborative learning setting 2 will be shown in the Appendix for conciseness.

\subsection{Collaborative Finetuning}\label{sec:exp_fl_finetune}

\begin{table}[h]
	\centering
	\caption{
		\textbf{Collaborative finetuning under IID setting (top) and non-IID setting 1 (bottom).} We \textit{finetune} the encoder and classifier with different ratios of locally labeled data on clients by supervised collaborative learning and report the top-1 accuracy.
	}
	\label{tab:exp_federated_finetune_iid_noniid1}
	\begin{minipage}{\linewidth}
		\setlength\tabcolsep{3.0pt}
		\renewcommand{\arraystretch}{1.0}
		\resizebox{1.0\columnwidth}{!}{
			\begin{tabular}{lccccccccc}
				\toprule
				& \multicolumn{2}{c}{CIFAR-10}    & \multicolumn{2}{c}{CIFAR-100}   & \multicolumn{2}{c}{FMNIST}      \\ \cmidrule(lr){2-3}\cmidrule(lr){4-5}\cmidrule(lr){6-7}
				Labeled ratio & 10\%           & 1\%            & 10\%           & 1\%            & 10\%           & 1\%            \\ \midrule
				FedRot        & 85.16          & 74.25            & 49.97               & 16.65             & 90.49               & 82.81          \\
				FedDC         & 79.98          & 71.17            & 42.81               & 21.47             & 90.17               & 84.22          \\
				FedSwAV       & 85.23          & \underline{78.92}            & 51.67               & \underline{26.75}             & 91.33               & \underline{85.10}          \\
				FedSimCLR     & \underline{83.52}          & 75.10            & \underline{51.73}               & 15.32             & \underline{91.64}               & 84.31          \\
				FedCA         & 82.32          & 72.77            & 50.78               & 21.10             & 91.57               & 84.29          \\
				Proposed      & \textbf{89.33}          & \textbf{82.52}            & \textbf{56.88}               & \textbf{33.15}             & \textbf{92.15}               & \textbf{87.11}          \\ \midrule
				FedAvg (Supervised)    & 74.71          & 39.35            & 33.16               & 8.07             & 87.95               & 75.68          \\
				\bottomrule
			\end{tabular}
		}
	\end{minipage}
	\begin{minipage}{\linewidth}
		\setlength\tabcolsep{3.0pt}
		\renewcommand{\arraystretch}{1.0}
		\resizebox{1.0\columnwidth}{!}{
			\begin{tabular}{lccccccccc}
				\toprule
				& \multicolumn{2}{c}{CIFAR-10}    & \multicolumn{2}{c}{CIFAR-100}   & \multicolumn{2}{c}{FMNIST}      \\ \cmidrule(lr){2-3}\cmidrule(lr){4-5}\cmidrule(lr){6-7}
				Labeled ratio & 10\%           & 1\%            & 10\%           & 1\%            & 10\%           & 1\%            \\ \midrule
				FedRot        & 57.34          & 56.80       & 46.93      & 17.12      & 75.02               & 72.02          \\
				FedDC         & 60.37          & 49.28       & 40.29      & 21.20      & 77.12               & 73.16          \\
				FedSwAV       & 57.34          & 51.93       & 46.65      & \underline{22.06}      & 75.45               & 73.17          \\
				FedSimCLR     & \underline{63.05}          & 51.63       & 47.69      & 11.19      & \underline{76.49}               & \underline{73.25}          \\
				FedCA         & 59.52          & \underline{57.33}       & \underline{49.14}      & 21.50      & 73.23               & 71.34          \\
				Proposed      & \textbf{65.80}          & \textbf{59.30}       & \textbf{50.75}      & \textbf{28.25}      & \textbf{78.81}             & \textbf{76.88} \\ \midrule
				FedAvg (Supervised)    & 48.41          & 32.33       & 33.26      & 8.42       & 67.42               & 66.03          \\
				\bottomrule
			\end{tabular}
		}
	\end{minipage}
\end{table}

We evaluate the performance of the proposed approaches by collaborative finetuning the learned encoder with few locally labeled data on clients.
The results under the IID setting and non-IID setting 1 are shown in Table \ref{tab:exp_federated_finetune_iid_noniid1} (top) and Table \ref{tab:exp_federated_finetune_iid_noniid1} (bottom), respectively. On both collaborative learning settings and three datasets, the proposed approaches consistently outperform the baselines.

\subsection{Ablations}

\begin{figure}[!htb]
	\centering
	\begin{subfigure}[b]{0.49\linewidth}
		\includegraphics[width=\linewidth]{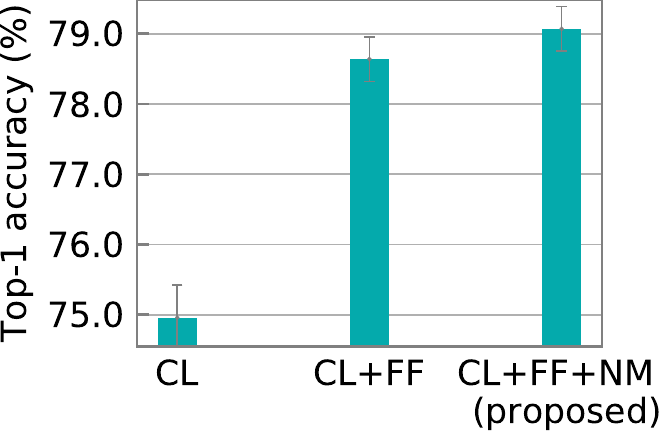}
		\caption{Linear evaluation.}
		\label{fig:}
	\end{subfigure}
	\begin{subfigure}[b]{0.49\linewidth}
		\includegraphics[width=\linewidth]{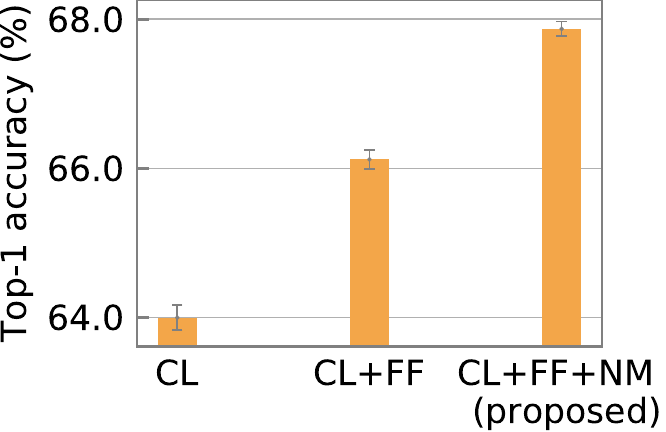}
		\caption{Semi-sup. learning.}
		\label{fig:}
	\end{subfigure}
	\caption{Ablations on CIFAR-10 dataset under the non-IID setting.
	CL is vanilla contrastive learning. FF is feature fusion and NM is neighborhood matching. Top-1 accuracy of linear classifier and semi-supervised learning (1\% labels) are reported.
	Error bar denotes standard derivation over three independent runs.}
	\label{fig:ablation_fe_nm_non_iid}
\end{figure}

\textbf{Effectiveness of feature fusion and neighborhood matching.}
We evaluate three approaches. 
Contrastive learning (CL) is the approach without feature fusion (FF) or neighborhood matching (NM).
CL+FF adds feature fusion, and CL+FF+NM further adds neighborhood matching.
We evaluate the approaches by linear evaluation and semi-supervised learning (1\% labels) under the non-IID collaborative learning setting (non-IID setting 1).
As shown in Figure \ref{fig:ablation_fe_nm_non_iid},
with linear evaluation, CL achieves 74.96\% top-1 accuracy. 
Adding FF improves the accuracy by 3.68\%, and adding NM further improves the accuracy by 0.43\%.
With semi-supervised learning (1\% labels), CL achieves 64.00\% top-1 accuracy.
Adding FF improves the accuracy by 2.12\% and adding NM further improves the accuracy by 1.75\%.
These results show the effectiveness of feature fusion and neighborhood matching.

\section{Conclusion}\label{sec:conclusion_and_limitation}

We propose a framework for collaborative contrastive representation learning. 
To improve representation learning on each client,
we propose feature fusion to provide remote features as accurate contrastive data to each client. 
To achieve unified representations among clients, we propose neighborhood matching to align each client's local features to the remote ones.
Experiments show superior accuracy of the proposed framework compared with the state-of-the-art.

\bibliographystyle{named}
\bibliography{ijcai22}

\end{document}